\documentclass[conference,onecolumn]{IEEEtran}

\usepackage{cite}
\usepackage[pdftex]{graphicx}
\graphicspath{{figures/}}
\usepackage[cmex10]{amsmath}
\usepackage{algorithmic}
\usepackage{algorithm}
\usepackage{array}
\usepackage{fixltx2e}
\usepackage{stfloats}
\usepackage{url}
\usepackage{amssymb}
\usepackage{float}
\usepackage{color}
\usepackage{balance}
\usepackage{subfig}
\usepackage{multirow}
\usepackage[margin=1in]{geometry}
\usepackage{placeins} %enables float barrier
\newcommand{\norm}[1]{\left\lVert #1 \right\rVert}

\begin{document}

\title{Training Stacked Denoising Autoencoders for Representation Learning}

\author{\IEEEauthorblockN{Jason Liang}
\IEEEauthorblockA{jasonzliang@utexas.edu}
\and
\IEEEauthorblockN{Keith Kelly}
\IEEEauthorblockA{keith@ices.utexas.edu}}

\maketitle

\begin{abstract}
We implement stacked denoising autoencoders, a class of neural networks that are capable of learning powerful representations of high dimensional data. We describe stochastic gradient descent for unsupervised training of autoencoders, as well as a novel genetic algorithm based approach that makes use of gradient information. We analyze the performance of both optimization algorithms and also the representation learning ability of the autoencoder when it is trained on standard image classification datasets. 
\end{abstract}

\FloatBarrier
\section{Introduction}
\begin{figure}[h]
\centering
\includegraphics[width=0.8\linewidth]{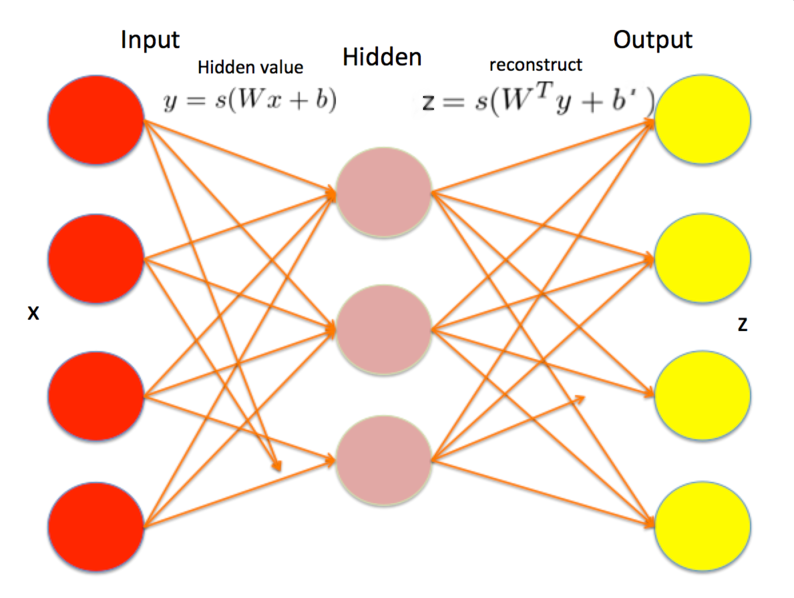}
\caption{Overview of an autoencoder and its encoding, decoding stages. The weight matrix of the decoding stage is the transpose of the weight matrix of the encoding stage.}
\label{fig:autoencoder}
\end{figure}

Autoencoders are a method for performing representation learning, an unsupervised pretraining process during which a more useful representation of the
input data is automatically determined. Representation learning is important in machine learning since ``the performance of 
machine learning methods is heavily dependent on the choice of data representation (or features) in which they are applied" 
\cite{bengio2012rep}. For many supervised classification tasks, the high dimensionality of the input data means that the classifier requires an enormous number of training examples in order to generalize well and not overfit. One solution is to use unsupervised pretraining to learn a good representation for the input data and during actual training, transform the input examples into an easier form for the classifier to learn. Autoencoders are one such representation learning tool.

An autoencoder is a neural network with a single hidden layer and where the output layer and the input layer have the same size. Suppose that the input $x\in\mathbb{R}^m$ (and the output as well) and suppose that the hidden layer has $n$ nodes. Then we have a weight matrix $W\in\mathbb{R}^{m\times n}$ and bias vectors $b$ and $b^{'}$ in $\mathbb{R}^m$ and $\mathbb{R}^n$, respectively. Let $s(x) = 1/(1+e^{-x})$ be the sigmoid (logistic) transfer function. Then we have a neural network as shown in Fig.~\ref{fig:autoencoder}. When using an autoencoder to encode data, we calculate the vector $y=s(Wx + b)$; corresponding when we use an autoencoder to decode and reconstruct back the original input, we calculate $z=s(W^{T}x+b^{'})$. The weight matrix of the decoding stage is the transpose of weight matrix of the encoding stage in order to reduce the number of parameters to learn. We want to optimize $W$, $b$, and $b^{'}$ so that the reconstruction is as similar to the original input as possible with respect to some loss function. In this report, the loss function used is the least squares loss: $E(t,z)=\frac{1}{2}\norm{t-z}_2^2$, where $t$ is the original input. After an autoencoder is trained, its decoding stage is discarded and the encoding stage is used to transform the training input examples as a preprocessing step. We will refer to the trained encoding stage of the autoencoder as an ``autoencoder layer".

Once an autoencoder layer has been trained, a second autoencoder can be trained using the output of the first autoencoder layer. This procedure can be repeated indefinitely and create stacked autoencoder layers of arbitrary depth. It is been shown that each subsequent trained layer learns a better representation of the output of the previous layer \cite{vincent2010stacked}. Using deep neural networks such as stacked autoencoders to do representation learning is also called deep learning, a subfield of machine learning that has received much attention and breakthroughs lately. 

For ordinary autoencoders, we usually want that $n<m$ so that the learned representation of the input exists in a lower dimensional space than the input. This is done to ensure that the autoencoder does not learn a trivial identity transformation. However, there also exists an autoencoder variant called \textit{denoising autoencoders} that use a different reconstruction criterion to learn overcomplete representations \cite{vincent2010stacked}. In other words, even if $n>m$, a denoising autoencoder can still learn a good representation of the input. This is achieved by corrupting the input image and training the autoencoder to reconstruct the original uncorrupted image. By learning how to denoise, the autoencoder is forced to understand the true structure of input data and learn a good representation of it. Although the loss function $E(t,z)$ for neural networks in general is non-convex, past work has shown that stochastic gradient descent (SGD) is sufficient for most problems. In this report, we will consider training denoising autoencoders with SGD. 

In addition, we will examine training autoencoders with other optimization methods such as a genetic algorithm (GA). A GA is a biologically inspired black-box optimization algorithm is capable of optimizing arbitrary non-convex, non-differentiable objective functions. The motivations behind using GAs for training autoencoders are two fold: 1) GAs are a novel approach to optimization of deep neural networks with large number of parameters such as autoencoders. We would like to evaluate the performance of GAs and compare it to SGD. 2) GAs have some advantages over SGD such as being able to escape local optima and easier to parallelize. However GAs also have drawbacks, the main one being its computational complexity over gradient descent. Unlike SGD, a GA must keep a population of individuals and must evaluate each individual for each training example. The computational complexity of a GA is $O(mnd)$ where $d$ is the dimensionality of the individual and optimization problem being solved, $n$ is the population size, and $m$ is the number of training examples; in comparison, the complexity of SGD is $O(md)$. The optimal population size depends on the problem being solved, but in most cases, $n=O(d)$. 

We will not just implement a conventional GA (CGA), but also explore hybrid GAs that also make use of gradient information (HGA). The key idea is that with additional gradient information to intelligently update the population every generation, the population size can be kept constant. As a result, the complexity of HGA becomes the same as SGD: $O(md)$, while still retaining the advantages of being more scalable and capable of optimizing arbitrary objective functions.

The rest of this report is as follows: Past research and background literature for autoencoders, dimensionality reduction, genetic algorithms, representation and deep learning can be found in the related work section. The algorithm description section contains more details about how SGD, CGA, and HGA are implemented for autoencoders. The experiments section describes the performance of SGA, CGA, and HGA, as well as the results of training an autoencoder on a handwritten digit image dataset. In the discussion section, we will analyze our findings and report key findings. Finally, in future work, we discuss possible areas of future research.

\FloatBarrier
\section{Related Work}
Dimensionality reduction of high dimensional data is a common problem in machine learning with many applications. Given input points embedded in high dimensional space, the goal of dimensionality reduction is to learn a lower dimensional manifold that fits most of the points well. Simpler methods such as principal component analysis (PCA) \cite{wold1987principal} work by finding the orthogonal directions that best explains the variance of the input data. These directions or ``principal components'' can be then used to transform the input data by mapping it to coordinates along the principal components. The main drawback of PCA is that it only learn a linear transformation, which means PCA performs poorly when the input data do not lie on a linear manifold (a hyperplane). More sophisticated variants such as kernel PCA \cite{scholkopf1997kernel} deal with nonlinear input by using a nonlinear transformation to map the input data points into a higher dimensional space where they are approximately linear. Other nonlinear manifold learning algorithms include Isomap \cite{tenenbaum2000global} and t-SNE \cite{van2008visualizing}. These algorithms use heuristics to compute a similarity metric between the input points and then perform multidimensional scaling. Unfortunately, all of these methods mentioned above only work well if the manifold is locally smooth and often degrade in performance for highly varying manifolds \cite{van2008visualizing}.

Autoencoders, also known as auto-associators, are a class of feed-forward neural networks that are designed for performing dimensionality reduction and manifold learning \cite{hinton2006reducing}. Artificial neural networks are biologically inspired computational structures that are capable of approximating functions of arbitrary shape and dimensionality \cite{haykin2004comprehensive}. Unlike normal neural networks for regression or classification, autoencoders are trained in an unsupervised manner. The weights of an autoencoder can be trained with any optimization algorithm, but the mostly commonly used one is gradient descent, also known as the backpropagation algorithm \cite{hecht1989theory,bottou-91c}. 

Most recent research on autoencoders focus on training them to perform representation learning \cite{bengio2012rep}. Representation learning is a generalization of manifold learning where the goal is not only to discover a lower dimensional substructure, but also to learn a higher dimensional but sparser and more linearly separable representation of the input data. It has been shown that training autoencoders to denoise is one way to learn good representations of the data \cite{vincent2010stacked}. Similarly, adding a regularization term that penalizes the Frobenius norm of the Jacobian matrix of the encoder activations is also effective \cite{rifai2011contractive}. Furthermore, by stacking autoencoders, it has been shown that the resulting deep neural network is able to overcome the difficulties associated with existing dimensionality reduction methods in learning highly varying manifolds \cite{van2008visualizing,vincent2010stacked}.

A related research area to representation learning is deep learning, which explores methods for training deep neural networks such that each subsequent layer learns a higher level representation of the previous layer's output \cite{bengio2009learning}. It has been shown with each additional layer, the transformed input feature steadily becomes more global and less invariant to local distortions. Compared to shallow neural networks, deep neural networks perform and generalize better on classification tasks, especially for high dimensional inputs. In fact for many difficult datasets, their performance is state of the art by a wide margin \cite{schmidhuber2014deep}. 

Besides stacked autoencoders, the other two major deep learning architectures are stacked restricted Boltzmann machines (RBMs) \cite{hinton2006fast} and convolutional neural networks (CNNs) \cite{lecun1995convolutional}. A RBM is an undirected graphical model that is trained to learn the distribution of the input data. The special bipartite structure of the RBM makes Gibbs sampling tractable and gives rise to a fast training algorithm called contrastive divergence \cite{hinton2006fast,hinton2006learning}. Like autoencoders, RBMs can also be stacked in a similar manner and converted into a deterministic deep neural network \cite{hinton2006reducing}. CNNs are a type of deep neural network that is especially designed for processing images and are composed of two main types of layers: convolutional and max pooling. In the convolutional layer, local filters of weights are convolved with the input while in the max pooling layer, the input is down-sampled. CNNs are trained in a supervised manner with backpropagation. Due to the sparsity of the architecture, CNNs can be efficiently trained for input data with millions of dimensions and consequently, achieve state of the art performance in many large scale image classification benchmarks \cite{ciresan2011committee,krizhevsky2012imagenet,goodfellow2013multi}.

A survey of GAs and how they perform optimization can be found in \cite{srinivas1994genetic}. Much research has be done in applying GAs for optimization of weights in neural networks \cite{gomez2006efficient,floreano2008neuroevolution}. There is a small amount of recent work on applying GAs and evolutionary algorithms in general to deep learning \cite{koutnik2014evolving,david2014genetic}. Cant{\'u}-Paz discusses methods for parallelizing GAs, which will be essential to scalability when using a GA to train our autoencoder \cite{cantu1998survey}.

\FloatBarrier
\section{Algorithm Description}
\subsection{Stochastic Gradient Descent}

We start with a random weight matrix $W$ and random biases $b$ and $b^{'}$. We take a given input $x$, and feed it forward through the network and compute the error between the target output $t$ and the actual output $z$. Often we use the squared loss error $E(t,z) = \frac{1}{2}\norm{t-z}_2^2$ to determine the difference between the two. In the case of an autoencoder, the target output is the same as the input. If the error is not satisfactory, we can adjust the weight matrix and that biases in order to attempt to learn a better representation of the data. A common method of updating the weight and biases is via backpropagation \cite{hecht1989theory}; when applied to training inputs one at a time, it is also known as stochastic gradient descent (SGD). We will first consider the update for the weights and biases from the last hidden layer to the output layer with a squared loss error function and derive the updates. We use as an example a simple three layer neural network (input, one hidden and output layer). Some notation is given in Table \ref{tab:notation}.

\begin{table}[h]
	\centering
\begin{tabular}{l|l}
Symbol     & Meaning                                                          \\ \hline
$E$          & Error as computed at the output layer                            \\
$x_j$       & Node $j$ in the input layer                                      \\
$y_j$       & Node $j$ in the hidden layer                                     \\
$z_j$       & Node $j$ in the output layer                                     \\
$n_j$       & $\sum_{i=1}^n W_{ij}x_i + b_j$                                   \\
$t_j$       & Target output at node $j$                                   \\
$W_{ij}^H$ & Weight $i,j$ from input to hidden layer \\
$W_{ij}^O$ & Weight $i,j$ from hidden to output layer \\
$s(x_j)$    & $1/(1+e^{-x_j})$                                                 \\
$b_j^{\{H,O\}}$    & Biases for hidden and output layer                                                \\
\end{tabular}
\caption{Table giving notation for the derivation of updates.}
\label{tab:notation}
\end{table}

The derivative of the output error $E$ with respect to an output matrix weight $W_{ij}^O$ is as follows.

\begin{equation}
\begin{split}
\frac{\partial E}{\partial W^O_{ij}} &= \frac{\partial E}{\partial z_j}\frac{\partial z_j}{\partial W^O_{ij}} \\
																	 &=(z_j - t_j)\frac{\partial s(n_j)}{\partial x_j}\frac{\partial x_j}{\partial W^O_{ij}} \\
																	 &=(z_j-t_j)s(n_j)(1-s(n_j))x_i \\
																	 &=(z_j-t_j)z_j(1-z_j)x_i \\
\end{split}
\label{}
\end{equation}
Now that we have the gradient for the error associated to a single training example, we can compute the updates.
\begin{equation}
\begin{split}
\delta^O_j &= (z_j-t_j)z_j(1-z_j) \\
W^O_{ij} &\leftarrow W^O_{ij} - \eta \delta^O_j x_i \\
b^O_j &\leftarrow b^O_j - \eta\delta^O_j
\end{split}
\end{equation}

The computation of the gradient for the weight matrix between hidden layers is similarly easy to compute.
\begin{equation}
\begin{split}
\frac{\partial E}{\partial W^H_{ij}} &= \frac{\partial E}{\partial y_j}\frac{\partial y_j}{\partial W^H_{ij}} \\
&=\left(\sum_{k=1}^m \frac{\partial E}{\partial z_k}\frac{\partial z_k}{\partial n_k}\frac{\partial n_k}{\partial y_j} \right)\frac{\partial y_j}{\partial n_j}\frac{\partial n_j}{\partial W_{ij}^H}\\
																	 &=\left(\sum_{k=1}^m (z_k - t_k)(1-z_k)z_kW_{jk}^O \right)y_j(1-y_j)x_i
\end{split}
\label{}
\end{equation}
And then using the computed gradient we can define the updates to be used for the hidden layers
\begin{equation}
\begin{split}
\delta^H_j &= \left(\sum_{k=1}^m (z_k - t_k)(1-z_k)z_kW_{jk}^O \right)y_j(1-y_j) \\
W^H_{ij} &\leftarrow W^H_{ij} - \eta\delta^H_jx_i \\
b^H_j &\leftarrow b^H_j - \eta\delta^H_j
\end{split}
\end{equation}

In general, for a neural network we may have different output error functions and these will result in different update rules. We will also give the updates
for the cross-entropy error function with softmax activation in the final layer.
The cross entropy error function is given by $E(x,t) = -\sum_{i=1}^n \left(t_i\ln z_i + (1-t_i)\ln(1-z_i)\right)$
and the softmax function is given by $\sigma(x_j) = e^{x_j} /(\sum_k e^{x_k})$. Following the same procedure as above
for computing the gradient and the updates, we find that for hidden/output layer

\begin{equation}
\begin{split}
\frac{\partial E}{\partial W^O_{ij}} &= (z_j - t_j)y_i \\
\delta^O_j &= (z_j-t_j) \\
W^O_{ij} &\leftarrow W^O_{ij} - \eta \delta^O_j x_i \\
b^O_j &\leftarrow b^O_j - \eta\delta^O_j.
\end{split}
\end{equation}

We find that the updates for the hidden layer is the same as in the squared error loss function with sigmoid activation. A general overview of the backpropagation algorithm is given by by Algorithm~\ref{alg:backprop}.

The algorithm and derivations for the autoencoder are a slight variation on the above derivations for a more general neural network. The weight matrix of the output layer (decoding stage) is the transpose of the weight matrix of the hidden layer (encoding stage). Thus $z=s(W^{O}(W^{H}x + b) + b^{'})$, $(W^H)^T = W^O$, and $W^H_{ij} = W^O_{ji}$. For training denoising autoencoders in particular, $z=s(W^{O}(W^{H}x_{\text{corr}} + b) + b^{'})$, where $x_{\text{corr}}$ is a randomly corrupted version of the original input data $x_{\text{orig}}$ and the loss function is defined as $E(x_{\text{orig}}, z)$. In order words, we are trying to learn an autoencoder takes in corrupted input and reconstructs the original uncorrupted version. Once we have trained a single autoencoder layer, we can stack another autoencoder layer on top of the first one for further training. This second autoencoder takes the corrupted output of the hidden layer (encoding stage) of the first autoencoder as input and is again trained to minimize the loss function $E(x_{\text{orig}}, z)$.

\begin{algorithm}[h]
\caption{Backpropagation}
\label{alg:backprop}
\begin{algorithmic}
\STATE Initialize the weights and biases randomly
\FOR{iter $=1,2,3...$}
	\FORALL{Examples $x$ in training set (randomize)}
		\STATE $z\gets$ Feedforward $x$
		\STATE Compute output layer $\delta_j^O$
		\STATE ${W_{ij} \leftarrow W_{ij} - \eta \delta^O_j x_i}$
		\STATE $b_j \leftarrow b_j - \eta\delta^O_j$
		\FORALL{Layers in reverse order}
			\STATE Compute hidden layer delta $\delta_k^H$
			\STATE ${W^H_{ij} \leftarrow W^H_{ij} - \eta\delta^H_jx_i}$
			\STATE $b_j \leftarrow b_j - \eta\delta^H_j$
		\ENDFOR
	\ENDFOR
\ENDFOR
\end{algorithmic}
\end{algorithm}

After using backpropagation (or a genetic algorithm) to train each of the autoencoder layers, we can then attach an output layer to the autoencoder to be used for classification. At this
point we use supervised learning to train the output layer and \textit{fine-tune} the autoencoder layers to produce a classifier based on the autoencoder. When pretraining the autoencoder
we train one layer at a time using backpropagation, but during the fine-tuning step we train the entire network via backpropagation, one layer at a time per training image.

\subsection{Genetic Algorithm}

As mentioned in the introduction, a genetic algorithm (GA) is a biologically inspired black-box optimization algorithm is capable of optimizing arbitrary non-convex, non-differential objective functions. The GA works by iteratively improving upon a population of candidate solution vectors (also know as individuals) via genetic operations such as selection, mutation, and crossover. The goal is to maximize the fitness of each individual, which is determined by evaluating the individual with some objective function. In the case of autoencoders, the objective function is the reconstruction loss function $E(t,z)$, the individual is a real valued vector that represents the weights ($W$) and biases ($b$, $b'$) and the fitness of an individual is simply $1/E(t,z)$. Algorithm~\ref{alg:genetic} shows the pseudocode of a simple conventional genetic algorithm, which we will refer to as CGA. 

\begin{algorithm}[h]
\caption{Conventional Genetic Algorithm (CGA)}
\label{alg:genetic}
\begin{algorithmic}
\STATE Initialize $N$ individuals randomly by uniformly sampling values from $[-r, r]$
\FOR{iter $=1,2,3,...,M$, where $M$ is number of training examples}
	\STATE Evaluate each individual with objective function and assign fitness
	\STATE Scale fitness of all individuals, typically with \textbf{power scaling} with parameter $\gamma$
	\STATE Select existing individuals via \textbf{roulette selection}
	\FOR{every two individuals $a$ and $b$ selected}
		\STATE Create copies $a'$ and $b'$
		\STATE Perform \textbf{uniform mutation} on $a'$ and $b'$ with mutation rate $mr$ and mutation amount $ma$
		\STATE Perform \textbf{uniform crossover} using $a'$ and $b'$ with crossover rate $cr$
	\ENDFOR
	\STATE Replace worst $\alpha N$ ($0< \alpha < 1$) individuals in population with newly created individuals
\ENDFOR
\end{algorithmic}
\end{algorithm}

We will explain step by step what each bold term means and how it affects individuals in the population:

\begin{itemize}
  \item Power Scaling: All individuals in the population are ranked in ascending order according to their fitness. Each individual is assigned a scaled fitness is is equal to $R^{\gamma}$ where R is the position of the individual within the ranking and $\gamma$ is a parameter to be tuned. Higher values for $\gamma$ will create a scaled fitness distribution that favor individuals with higher actual fitness. Thus $\gamma$ will affect the selection of individuals later and determine how elitist the selection of individuals will be (in other words, how much more likely an individual of higher fitness will be selected).
  \item Roulette Selection: Also known as fitness proportionate selection, roulette selection randomly samples an individual from the population with probability proportional to its scaled fitness. There are many other selection methods such as tournament selection and truncation selection. However, our preliminary experimental results suggest that roulette selection is most appropriate for training weights of autoencoders. The purpose of this operation is to select individuals with good fitness for creating the next generation's population. 
  \item Uniform Cauchy Mutation: Each element of the individual (a vector of real numbers) is mutated with probability $mr$. For each element, we generate a random number between 0 and 1. If this random number is less than $mr$, we sample a value from a zero-mean Cauchy distribution with standard deviation $ma$ and add that value to the element. The purpose of this operation is to randomly perturb existing individuals and possibly create new ones whose fitness is slightly better. 
  \item Uniform Crossover: Each element of the individual is swapped with the same corresponding element of another individual with probability $cr$. If crossover is performed on two individuals with good fitnesses, there is a chance that the elements of the individuals that are responsible for their high fitnesses might be combined together into a new individual. 
\end{itemize}

Next we will look at HGA, a hybrid GA that combines genetic operators with backpropagation to keep the population size small and remain computationally tractable. The pseudocode for HGA is shown in Algorithm~\ref{alg:hga} and is based off recent work done by David and Greental \cite{david2014genetic} on applying GAs to learn stacked autoencoders. The main difference between our HGA and the algorithm described in \cite{david2014genetic} is that we use roulette selection instead of truncation selection and uniform Cauchy mutation instead of mutating weights to zero.

\begin{algorithm}[h]
\caption{Hybrid Genetic Algorithm (HGA)}
\label{alg:hga}
\begin{algorithmic}
\STATE Initialize $N$ individuals randomly by uniformly sampling values from $[-r, r]$
\FOR{iter $=1,2,3,...,M$, where $M$ is number of training examples}
	\STATE Evaluate each individual with objective function and assign fitness
	\STATE Scale fitness of all individuals, typically with \textbf{power scaling} with parameter $\gamma$
	\STATE \textit{Perform backpropagation on top $\beta N$ ($0 < \beta < 1$) individuals with highest fitness}
	\STATE Select existing individuals via \textbf{roulette selection}
	\FOR{every two individuals $a$ and $b$ selected}
		\STATE Create copies $a'$ and $b'$
		\STATE Perform \textbf{uniform mutation} on $a'$ and $b'$ with mutation rate $mr$ and mutation amount $ma$
		\STATE Perform \textbf{uniform crossover} using $a'$ and $b'$ with crossover rate $cr$
	\ENDFOR
	\STATE Replace worst $\alpha N$ ($0< \alpha < 1$) individuals in population with newly created individuals
\ENDFOR
\end{algorithmic}
\end{algorithm}

The key idea of performing backpropagation is that it assists the mutation operator in moving the individuals towards regions of high fitness. While the mutation operator perturbs individuals randomly, backpropagation will always follow the gradient. This distinction is significant in high dimensional spaces, as the mutation operator will require a large population size to work optimally, while backpropagation does not. Instead, the mutation operator serves a secondary role of helping the individuals to escape from a local optima through random perturbations. Preliminary results show that compared to CGA with the same population size, HGA performs significantly better. Furthermore, the performance of HGA does not degrade significantly with smaller population sizes, even when using an extremely small population size of 2. 
\FloatBarrier

\section{Experimental Results}
\FloatBarrier

For the following experiments, we train our autoencoder over the MNIST handwritten digit
dataset. The MNIST dataset is composed of 60000 training images and 10000
testing images Each image is in greyscale, is 28 by 28 pixels in size, and has
a corresponding label ranging from 0 to 9. Thus, the input vector for our
autoencoder has 784 dimensions. We also make use of the denoising criterion
mentioned in \cite{vincent2010stacked}, and for each training image, randomly
corrupt it by setting each pixel to zero with probability 0.25. All experiments were ran on a TACC Stampede cluster with 16 cores. 

\subsection{Stochastic Gradient Descent}

\begin{figure}[h]
\centering
\includegraphics[width=0.8\linewidth]{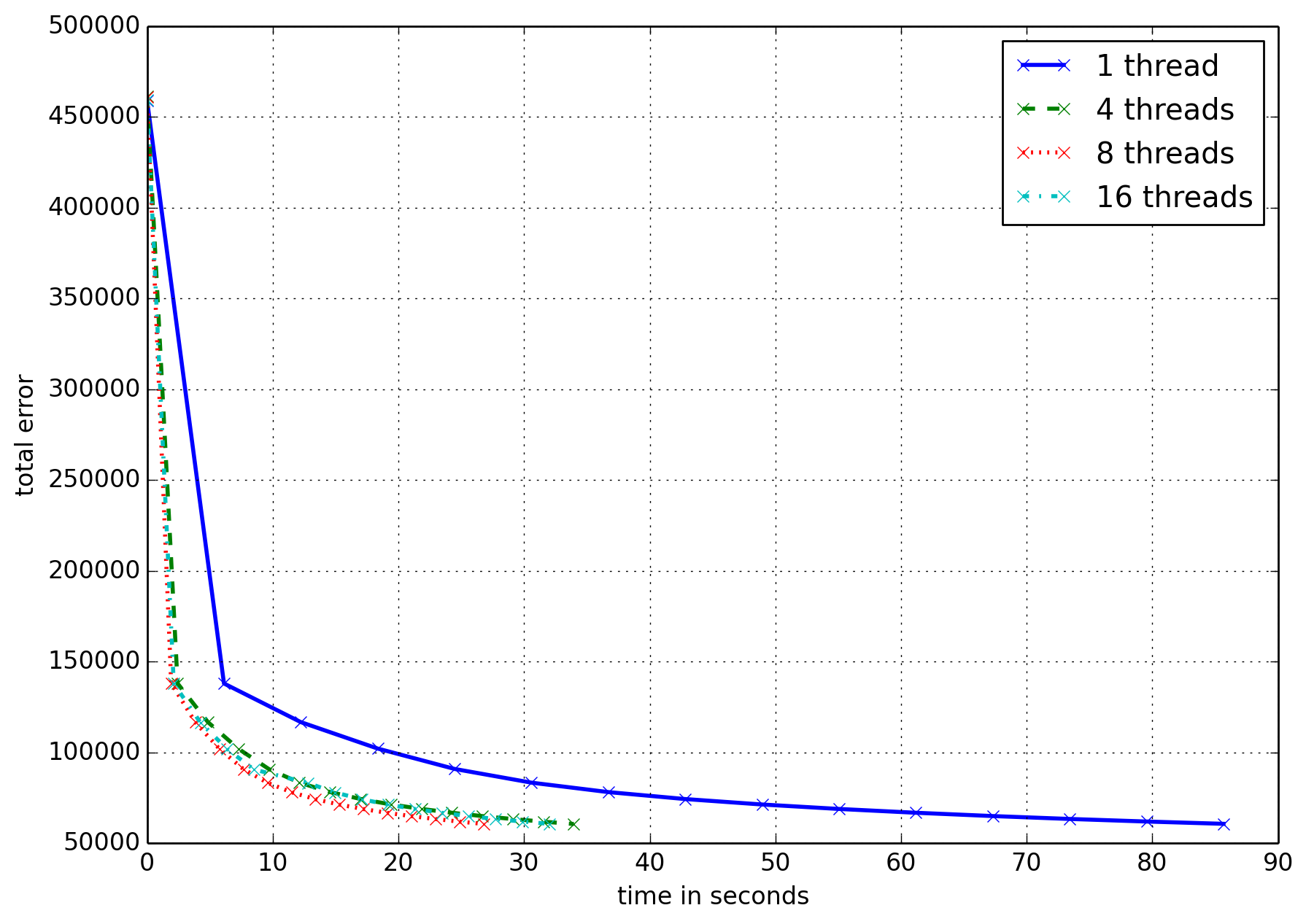}
\caption{Performance results on a single autoencoder layer with 500 hidden nodes and trained for 15 iterations. Plot shows time elapsed versus total training error over 5000 images for 1, 4, 8, and 16 threads.}
\label{fig:experiment1}
\end{figure}

We first analyze how the number of threads affects the rate at which reconstruction
error decreases for SGD. We train a single autoencoder layer with 500 hidden nodes for
15 iterations over 5000 training images. An iteration involves going through
all training images and for each image, use SGD to update the weight matrix. We initialize the weights of the autoencoder randomly by sampling from the interval $[-1/\sqrt{\text{FANIN}}, 1/\sqrt{\text{FANIN}}]$, where FANIN is the dimensionality of the input.
Fig.~\ref{fig:experiment1} shows the relationship between reconstruction error, total
time elapsed, and the number of threads used. Regardless of the number of
threads, the reconstruction error decreases sharply in the first few iterations
before flattening out to around the same value after 15 iterations. The rate at
which error decreases is significantly faster for 4 and 8 threads when
compared to just using one. Nonetheless, the speedup is not linear (using 16 threads is actually marginally slower than using 8) and is due
to two reasons: 1) Possible cache conflicts as each thread read and writes to
different locations in the weight matrix. 2) All the steps for
backpropagation/SGD, described in Algorithm \ref{alg:backprop}, must be done
sequentially. Parallelization can only be done within each step and incurs an
overhead cost.

We note that the steps of backpropagation can only be done in sequence; thus we can only parallelize the operations done within each step. The three major operations which benefit from parallelization are computing the matrix-vector products $W^{H}x$ and $W^{O}y$, computing $\delta^O_j$ and $\delta^H_j$, and updating the entries of the weight matrices with the gradient. For performance reasons we don't store $W^O$ separately; instead we access $W^H$ with transposed indexes when decoding, calculating $\delta^O_j$, and applying the gradient update.

The most expensive parts of the backpropagation algorithm are computing the forward activations of the network and updating the weight matrices for the network. Computing the forward propagation requires performing a matrix-vector multiplication at each layer of the network. The size of that matrix depends on the input and output sizes of that layer. Thus if we have a network with $N$ layers and the sizes of each of those layers is $n_i$, then We have $N$ matrix-vector multiplications of size $n_{i-1} \times n_i$. Updating the weights also has complexity based on the size of the matrix since it requires updating all the entries of the matrix at each iteration.

\begin{figure}[h]
\centering
\includegraphics[width=0.8\linewidth]{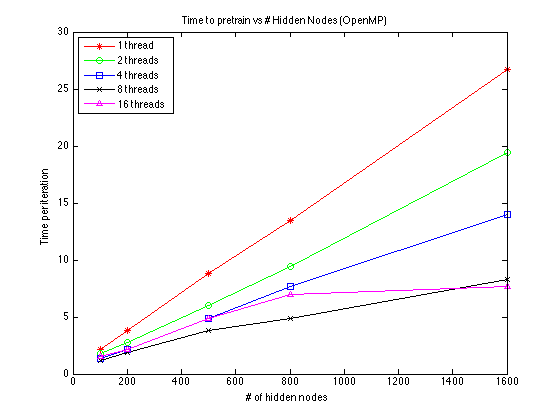}
\caption{Time per iteration versus the number of threads and hidden nodes. We parallelize with our own OpenMP implementation and use 5000 training images.}
\label{fig:performanceomp}
\end{figure}

\begin{figure}[h]
\centering
\includegraphics[width=0.8\linewidth]{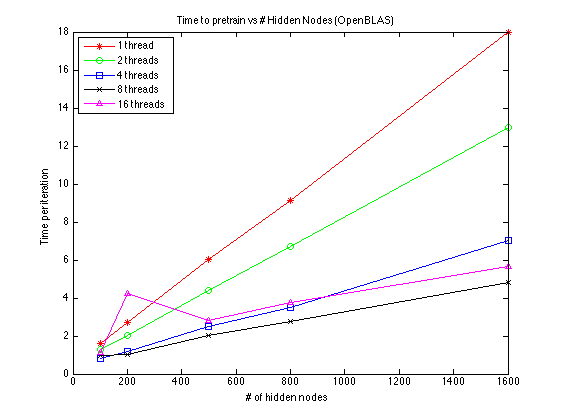}
\caption{Time per iteration versus the number of threads and hidden nodes. We parallelize with OpenBLAS and we use 5000 training images.}
\label{fig:performanceblas}
\end{figure}

\begin{figure}[h]
\centering
\includegraphics[width=0.8\linewidth]{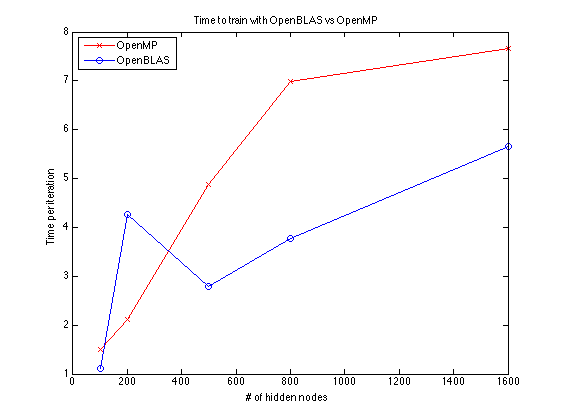}
\caption{Time per iteration versus parallelization technique and hidden nodes. We use 16 threads here.}
\label{fig:ompvsblas}
\end{figure}

To improve the performance of these two expensive steps, we parallelized them first using OpenMP and then later parallelized the matrix-vector products using OpenBLAS. Parallelization of these two steps is fairly straight forward and we give some results on scaling in Figs.~\ref{fig:performanceomp},~\ref{fig:performanceblas}, and \ref{fig:ompvsblas}.

In  Fig.~\ref{fig:performanceomp} we give the performance of training the autoencoder using SGD for different numbers of hidden nodes and different numbers of threads using our own parallel implementation.
In  Fig.~\ref{fig:performanceblas} we give the performance of training the autoencoder using SGD for different numbers of hidden nodes and different numbers of threads using OpenBLAS for parallelization of the matrix-vector multiplies and matrix-transpose-vector multiplies. The weight updates are not able to be done in OpenBLAS, and so we continue to do that using OpenMP. We consider the relative performance of these two methods in Fig.~\ref{fig:ompvsblas}.

We note that Figs.~\ref{fig:performanceomp} and~\ref{fig:performanceblas} show similar scaling, though in most cases the OpenBLAS version outperforms our own implementation. However, for small numbers of nodes, OpenBLAS does not perform well with many threads.
We generally get better performance with a greater number of threads, but with a large number of threads we do not see improvement until the problem size increases and becomes large enough. We do not achieve linear scaling (i.e. twice as many threads does not result in the algorithm running twice as fast), since not all parts of the algorithm can be parallelized. 

\subsection{Visualization of Trained Weights and Reconstructed Images}

\begin{figure}[h] \centering
  \includegraphics[width=0.8\linewidth]{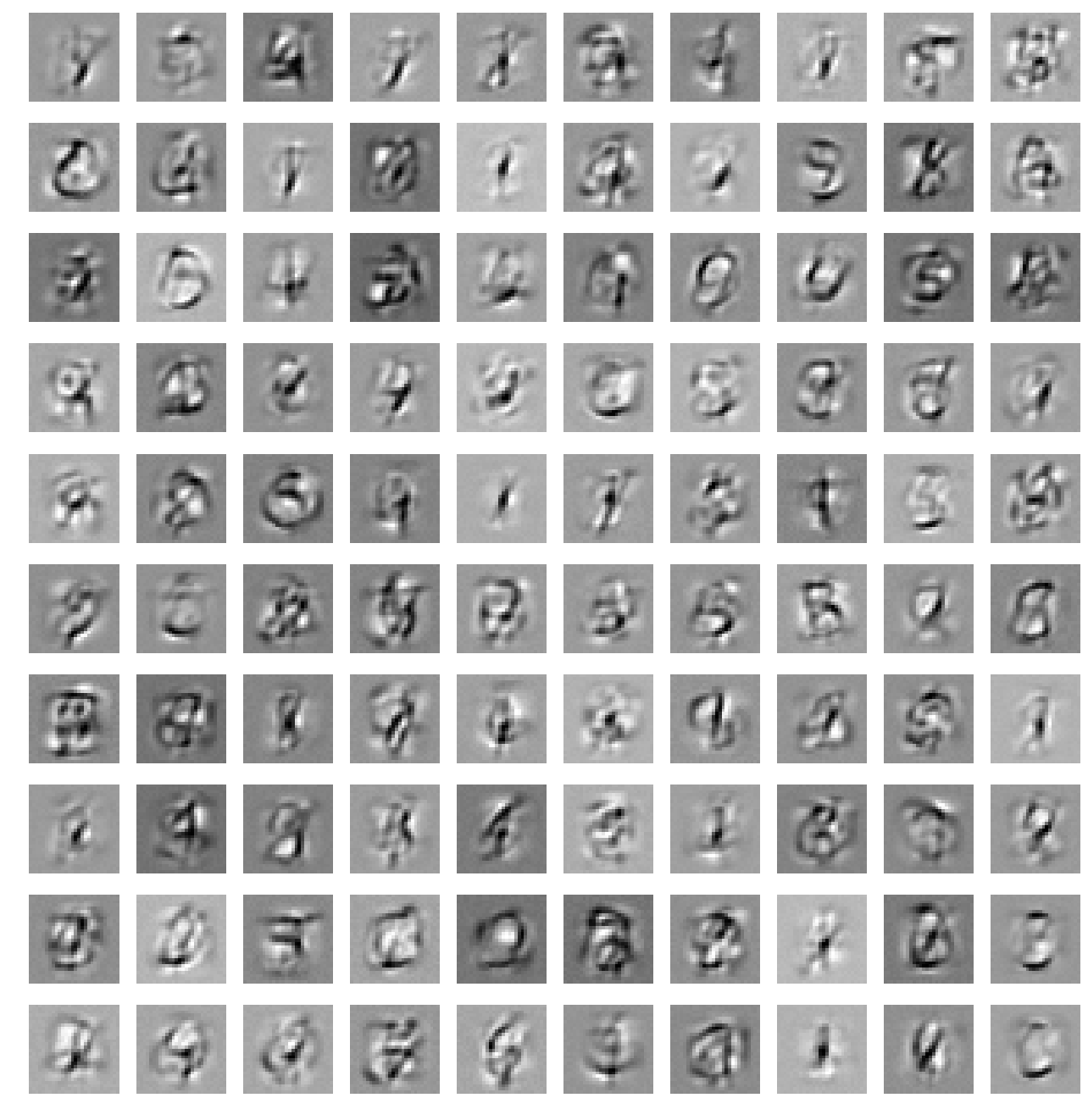}
  \caption{Visualization of the filters of the first 100 hidden nodes in an
  denoising autoencoder trained over all 60000 images.}
  \label{fig:experiment3_1}
\end{figure}

\begin{figure*}
  \centering
  \subfloat[]{
    \includegraphics[width=0.4\linewidth]{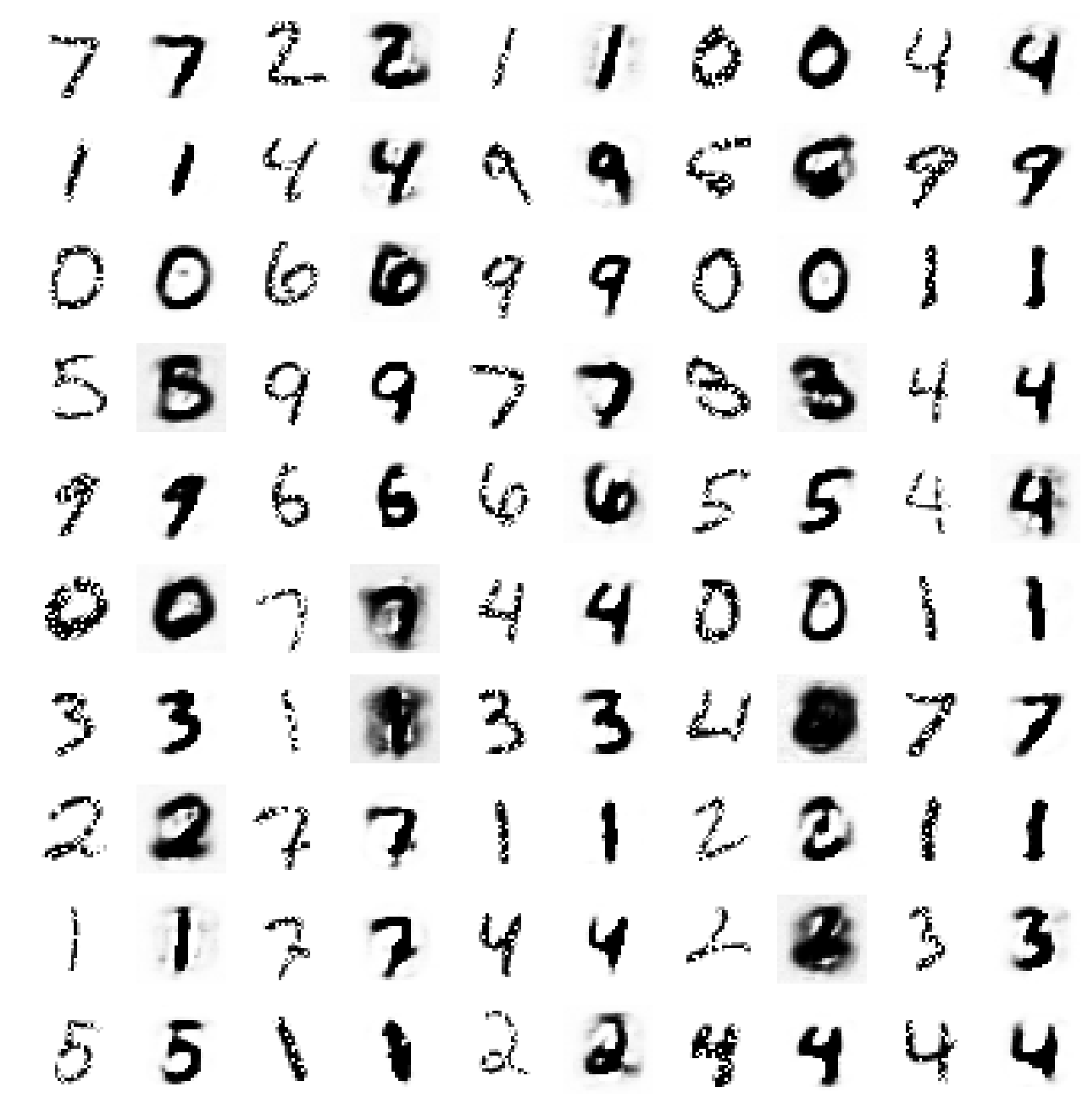}
  }
  \subfloat[]{
    \includegraphics[width=0.4\linewidth]{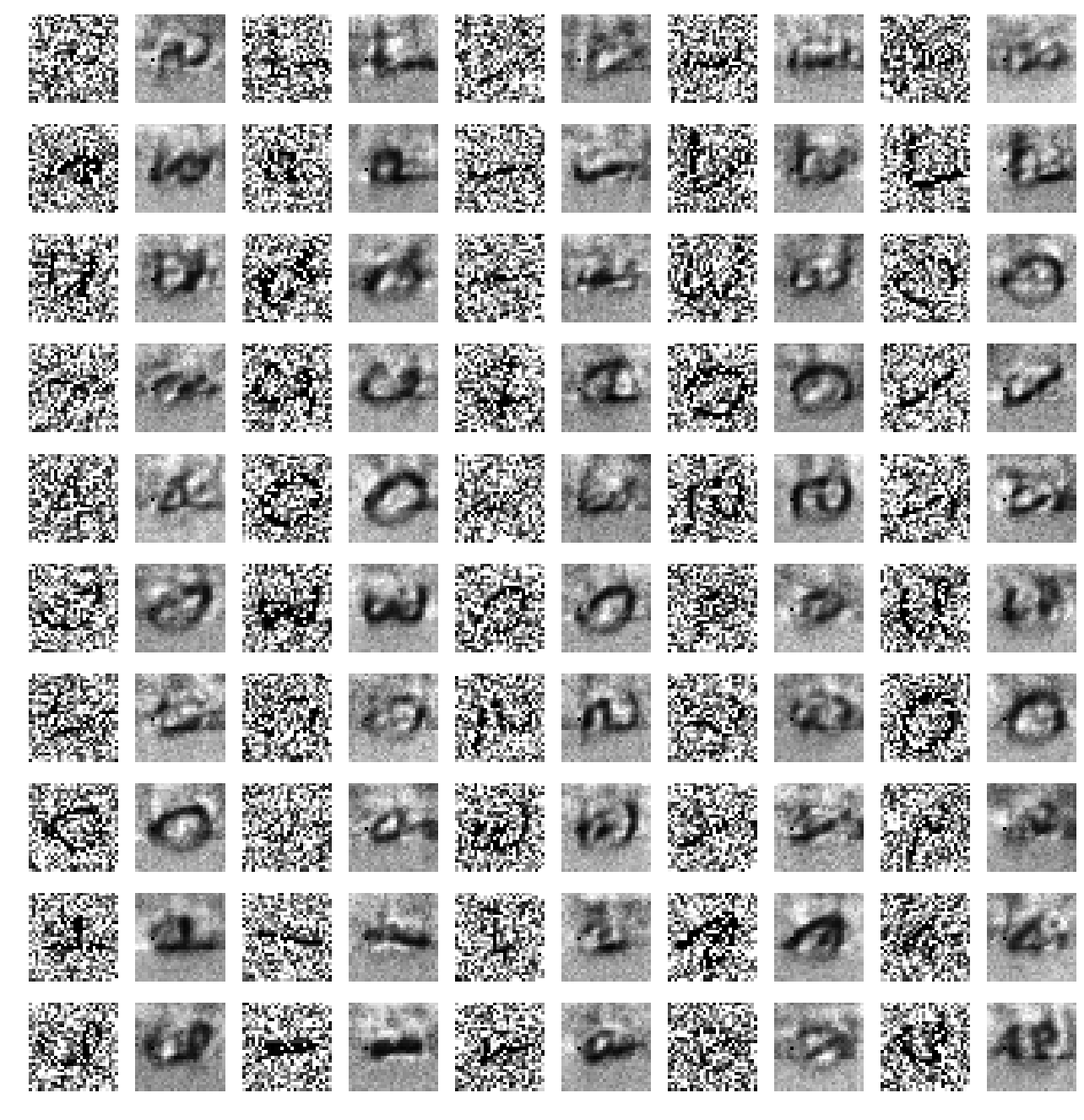}
  }
  \hspace{5mm}
  \subfloat[]{
    \includegraphics[width=0.4\linewidth]{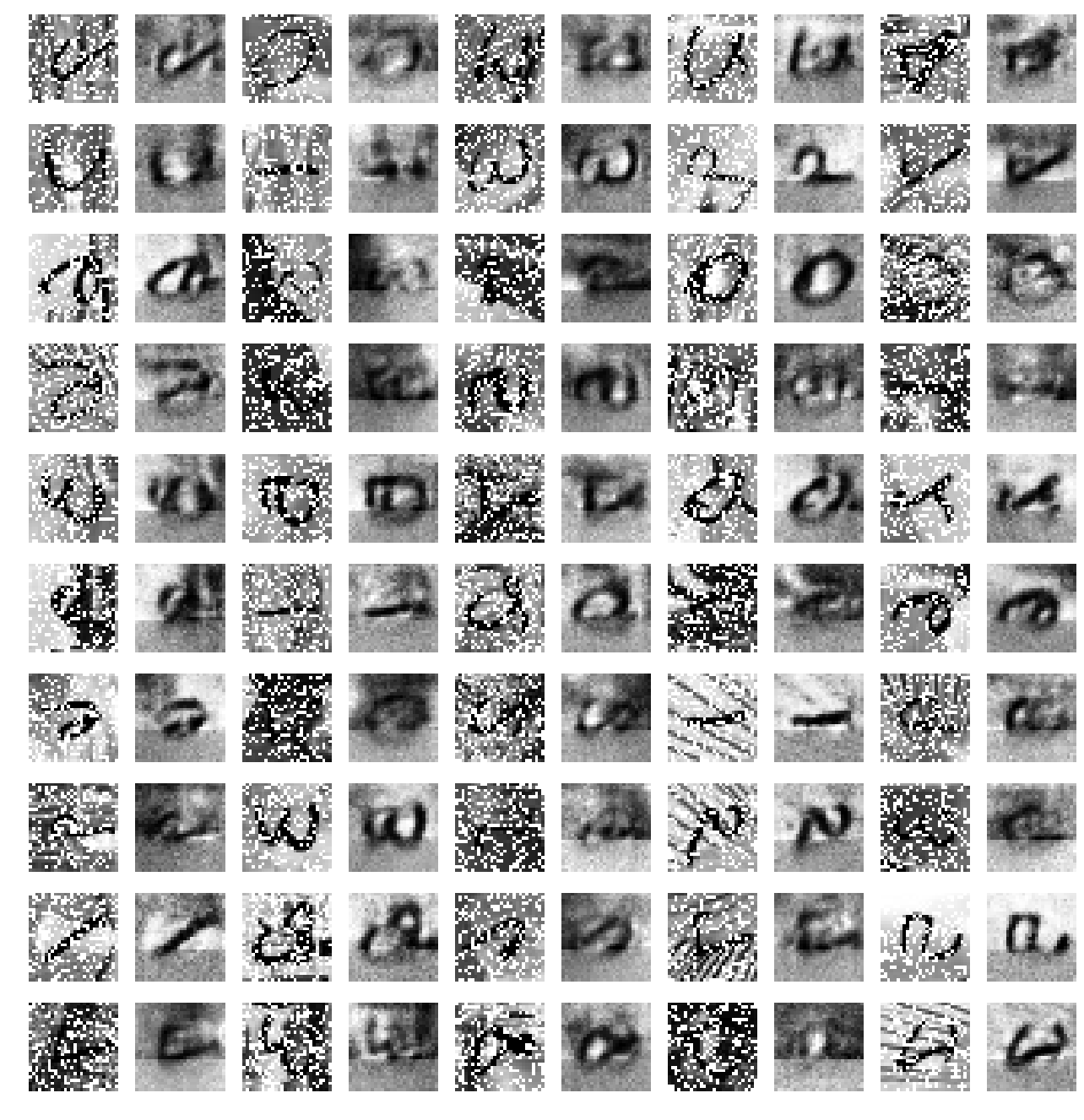}
  }
  \subfloat[]{
    \includegraphics[width=0.4\linewidth]{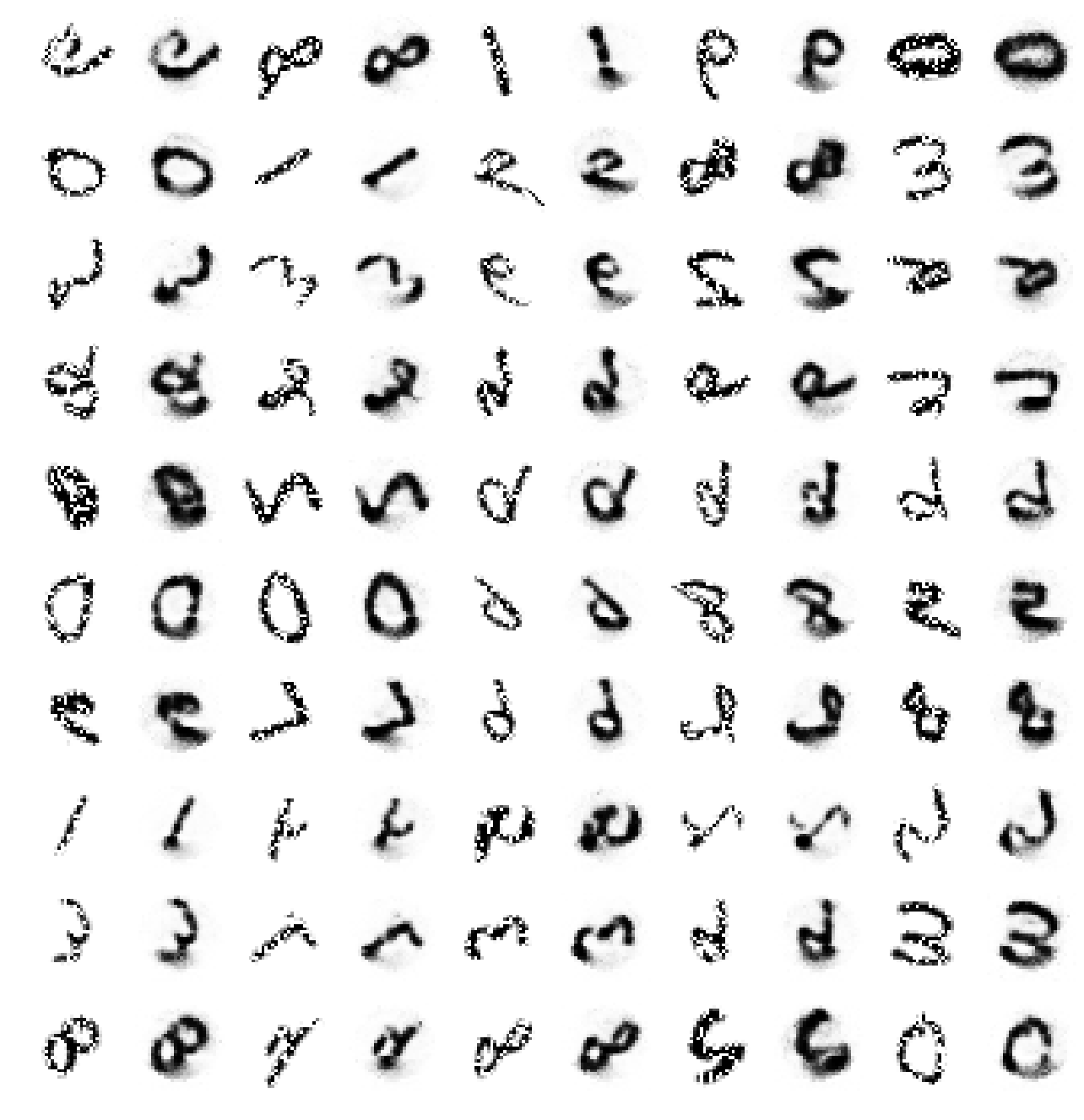}
  }
  \caption{Reconstructions of corrupted digits from the (a) MNIST dataset (b) bg-rand dataset, (c) bg-img dataset, and (d) rot dataset}
  \label{fig:reconstruct}
\end{figure*}

Next, in Fig.~\ref{fig:experiment3_1}, we visualize the filters that are
learned by training an autoencoder layer with 500 hidden nodes over all 60000
training images. The  filter for each hidden node is a row vector of the weight
matrix and indicates which aspects of the input the hidden unit is sensitive
to. Since each row in the weight matrix is the same dimensionality as the
input, we can visualize it as a 28 by 28 pixel image. The filters are not
identical to the input images, but do show some similarity to them. In
Fig.~\ref{fig:reconstruct}, we visualize the reconstructed digits when given
noisy test digits as input. The reconstructed outputs for most of the input
images are easily recognizable as digits, which indicates that the autoencoder
is indeed denoising and learning a good representation of the images.

To further demonstrate the reconstruction capabilities of the autoencoder, we
trained the autoencoder on corrupted or otherwise altered digits.  In
particular, we test the reconstruction on \textit{bg-rand}, which is generated
from the MNIST dataset, except that a random background is added. In
\textit{bg-img}, each image is given a background randomly selected from one of
twenty images downloaded from the internet. In \textit{rot}, the digits are
simply rotated by some random angle.  These alterations to the images make the
classification task more difficult. Indeed the digits are very difficult to
identify, but the autoencoder creates an easier to identify representation,
even to the human eye. We show also in Fig.~\ref{fig:reconstruct} the
reconstructed images form the \textit{bg-img} and \textit{rot} datasets. Note
that the images in the \textit{bg-rand} and \textit{bg-img} datasets are
rotated, but they are all rotated in the same way.

Finally we evaluate the classification accuracy of a deep neural network that
has multiple stacked denoising autoencoders. We train 3 stacked autoencoder
layers, each with 1000 hidden units, and using noise levels 0.1, 0.2, and 0.3
respectively. Each layer is trained for 15 iterations with a learning rate of
0.001. After the unsupervised pretraining, a conventional feed-forward network
with 1000 input units, 500 hidden units and 10 outputs is connected to the
hidden units of the last autoencoder layer. This conventional network is then
trained for 30 iterations (learning rate 0.1) in a supervised manner, where the
target $t$ is the indicator vector representation of the training label. Our
final classification accuracy is 98.04\%. In comparison, the best reported accuracy achieved
with a SVM with RBF kernel is 98.60\% \cite{vincent2010stacked}.

\subsection{Representation Learning for Supervised Classification}

Recall that one of the main reasons for using an autoencoder is to determine a
more useful representation of the data for other tasks, for example in a
classification task. To this end, we constructed and trained (15 iterations) an
autoencoder with just a single layer and 1000 hidden units and used it to
create a more useful representation of the digits in the MNIST dataset. After
this more useful representation is constructed, we can then use the output from
the autoencoder as input to another type of classification algorithm.  Since
the autoencoder produces a better representation of the data, we expect that
given the encoded data, the other classification algorithms should perform
better.  The results of these experiments is given in
Table.~\ref{tab:classvsenc}.

To test this, we used liblinear \cite{fan2008liblinear} to attempt to train a model and then predict on
a test set for both the original and encoded datasets. With the original data,
liblinear gives an accuracy of 91.68\% on the test set when using the default
parameters. However, the encoded data from the trained autoencoder gives
an accuracy of 97.07\%. This is a nontrivial improvement in the classification
accuracy. Thus, the autoencoder has created a better representation of the data
which made it easier for liblinear to classify. This verifies that the
autoencoder is doing what it is expected to do.

Similarly, we performed the same experiment as above, except in this case we
used libsvm \cite{chang2011libsvm} with an RBF kernel and all the default parameters.  Without
encoding the data first, we get an accuracy of 94.46\%, but using the encoded
data gives a prediction accuracy of 95.48\%. As above, the encoded data allows libsvm
better classify the data. We should note that our accuracy is not as good as the one reported earlier as we did not tune the hyperparameters (C and kernel width) of the SVM classifier. 

Using logistic regression to perform the classification, we experienced similar results.
Again we use liblinear with all default options except selecting logistic regression. Using the
original MNIST data, this algorithm achieved an accuracy of 91.82\% while with the encoded data
we achieved an accuracy of 96.86\%.

We performed the same experiments on the \textit{bg-rand}, \textit{bg-img} and 
\textit{rot} datasets as well. Across the board we see similar results. The encoding does not have much of
an effect on the ability of kernel SVM to classify the data, but for linear SVM and logistic regresssion, 
using the autoencoder always results in the classifier improving it's accuracy.

\begin{table}[h]
	\centering
\begin{tabular}{ll|lll}
    Dataset                        & Data               & Linear SVM & Kernel SVM (RBF) & Logistic Regression \\ \hline
    \multirow{2}{*}{MNIST}         & Original           & 91.68\%    & 94.46\%          & 91.82\%             \\
                                   & Encoded            & 97.07\%    & 95.48\%          & 96.86\%             \\ \hline
    \multirow{2}{*}{mnist-bg-rand} & Original           & 58.975\%   & 83.875\%         & 65.5917\%           \\
                                   & Encoded            & 81.675\%   & 83.6583\%        & 83.825\%            \\ \hline
    \multirow{2}{*}{mnist-bg-img}  & Original           & 69.25\%    & 76.4667\%        & 71.6333\%           \\
                                   & Encoded            & 78.4333\%  & 72.8417\%        & 78.1333\%           \\ \hline
    \multirow{2}{*}{mnist-rot}     & Original           & 11.204\%   & 13.328\%         & 11.868\%           \\
                                   & Encoded            & 18.428\%   & 14.378\%         & 18.78\%
\end{tabular}
	\caption{Summary of the results of running different classification algorithms on the raw MNIST data and on the output from a trained autoencoder. We
	see in nearly all cases that using the encoded data produces a better result.}
	\label{tab:classvsenc}
\end{table}

\subsection{Genetic Algorithms}

\begin{table}[h]
  \centering
\begin{tabular}{l|l}
Hyperparameter     & Value                                                          \\ \hline
Mutation Rate $mr$ & $0.0001$ \\
Mutation Amount $ma$ & $0.1r$ \\
Crossover Rate $mr$ & $0.5$ \\
Population Size $N$ & $2$ \\
Replacement Fraction $\alpha$ & $0.5$ \\
Initial Value Range $r$ & $1/\sqrt{\text{FANIN}}$ \\
Power Scaling Parameter $\gamma$ & $1.0$ \\
Backpropagation Fraction $\beta$ (HGA only) & $0.5$ \\ 

\end{tabular}
\caption{List of hyperparameters for CGA and HGA. FANIN is input dimensionality of autoencoder layer.}
\label{tab:hyperparameters}
\end{table}

\begin{figure}[h] \centering
  \includegraphics[width=0.8\linewidth]{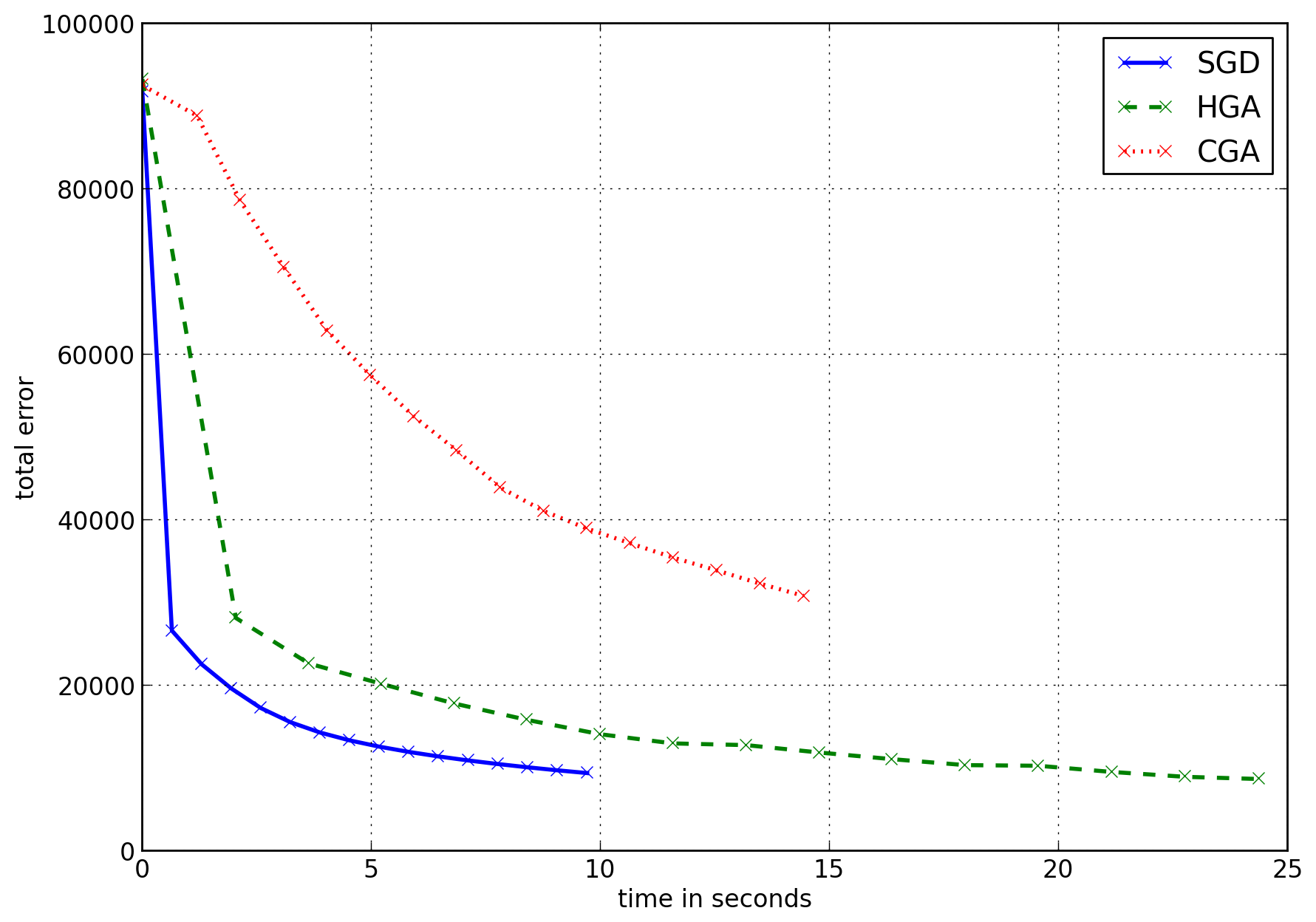}
  \caption{Comparison of the performance of SGD, HGA, and CGA. SGD is fastest, while HGA achieves the lowest reconstruction error. HGA and CGA uses a default population size of 2 and all three algorithms cycle through 1000 training images for 15 iterations.}
  \label{fig:ga_comparison}
\end{figure}

For all of the following experiments, we train a single autoencoder layer with 1000 hidden units and cycle through 1000 training digits for 15 iterations. The default hyperparameters for CGA and HGA are listed in Table.~\ref{tab:hyperparameters}. To parallelize the GA, we modified the mutation and crossover operators to support multiple threads. Because mutation and crossover modifies each element of the individual independently, no locking is required. The computation of an individual's fitness is already parallelized since it reuses the same code from SGD for determining the least squares loss. The remaining operations of the GA does not need parallelization since they are computationally cheap. 

In Fig.~\ref{fig:ga_comparison}, we compare the performance of SGD, HGA (uses backpropagation to update the best individuals in the population), and CGA (does not use any gradient information). SGD is the fastest by roughly a factor of two when compared to HGA, while CGA is somewhere in between. However, HGA is able to achieve the lowest reconstruction error (8709 vs SGD's 9427). CGA performed the worst out of all three algorithms, having a reconstruction error that is three times larger than that of the other algorithms. The hyperparameters for HGA and CGA are hand tuned and not necessary optimal. However, we believe that with properly tuned hyperparameters, HGA might be competitive with SGD for both final reconstruction error and performance time. 

\begin{figure}[h] \centering
  \includegraphics[width=0.8\linewidth]{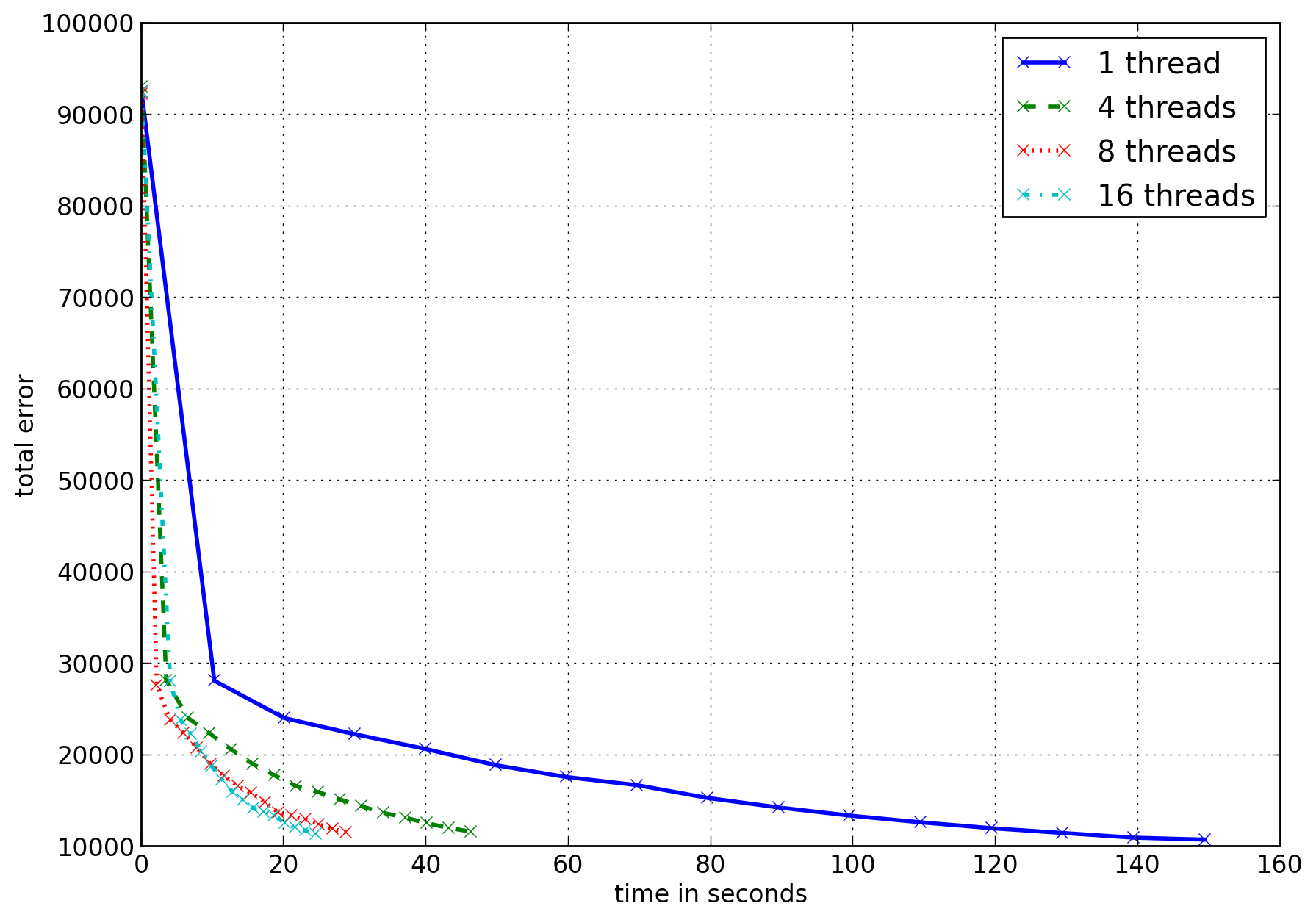}
  \caption{Performance of HGA for 1, 4, 8, and 16 threads with a population size of 2 and 1000 training images.}
  \label{fig:ga_comparison2}
\end{figure}

Next in Fig.~\ref{fig:ga_comparison2}, we examine the scalability of HGA. HGA shows roughly 3x speedup when using 4 threads, 5x speedup when using 8 threads, and 6x speedup when using 16 threads. The performance improvement versus the number of threads is sub-linear and can be attributed to two main causes: 1) Cache conflicts when performing mutation and crossover due to multiple threads writing and reading different memory locations, 2) The sequential nature of backpropagation, which HGA utilizes to help optimize the weights. However, HGA does seem to have moderately better scalability than our parallel implementation of SGD. This is attributable to the fact that the mutation and crossover operators are very straightforward to parallelize and scale in performance very well. 

\begin{figure*}[h]
  \centering
  \subfloat[]{
    \includegraphics[width=0.45\linewidth]{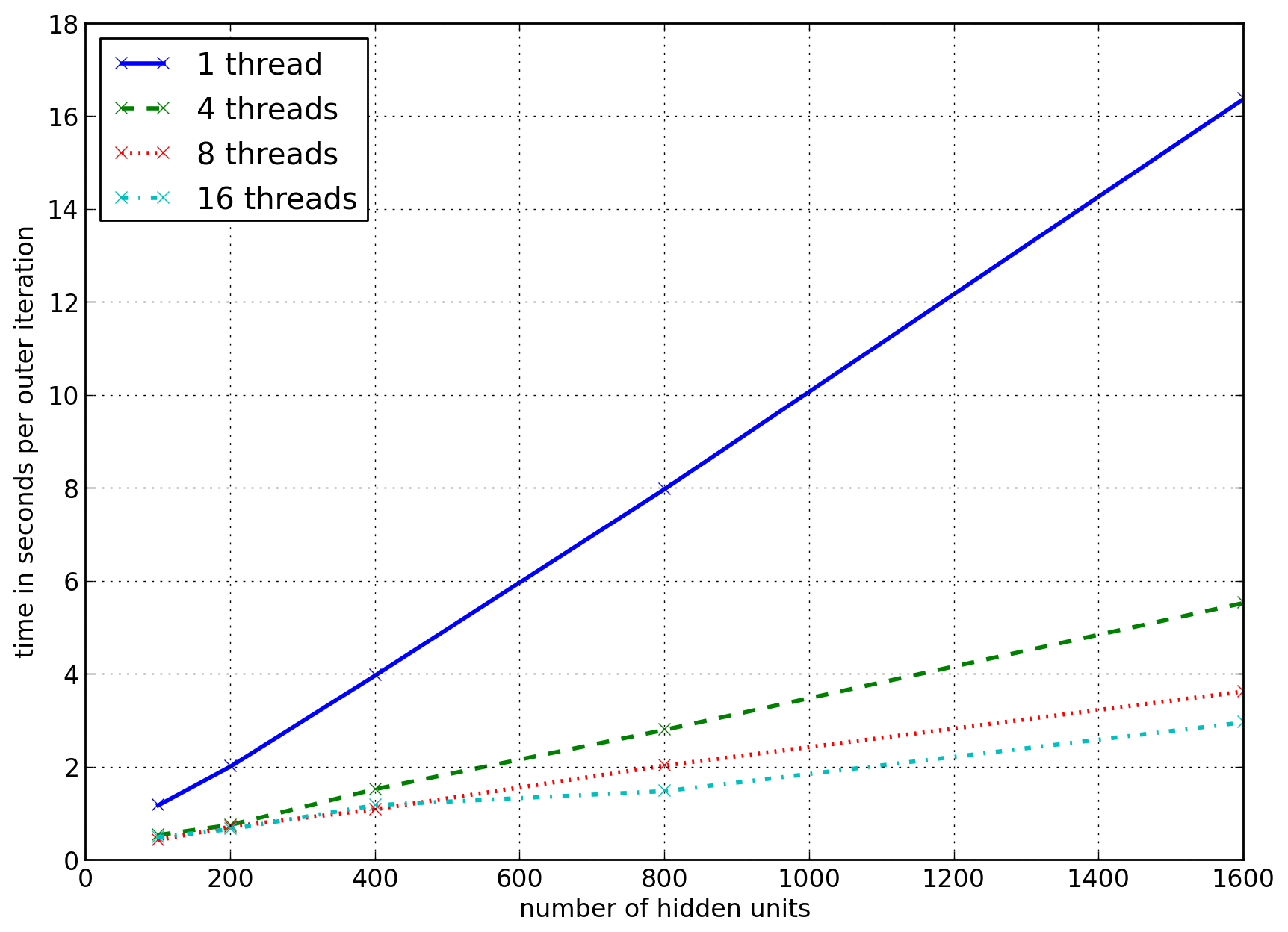}
  }
  \subfloat[]{
    \includegraphics[width=0.45\linewidth]{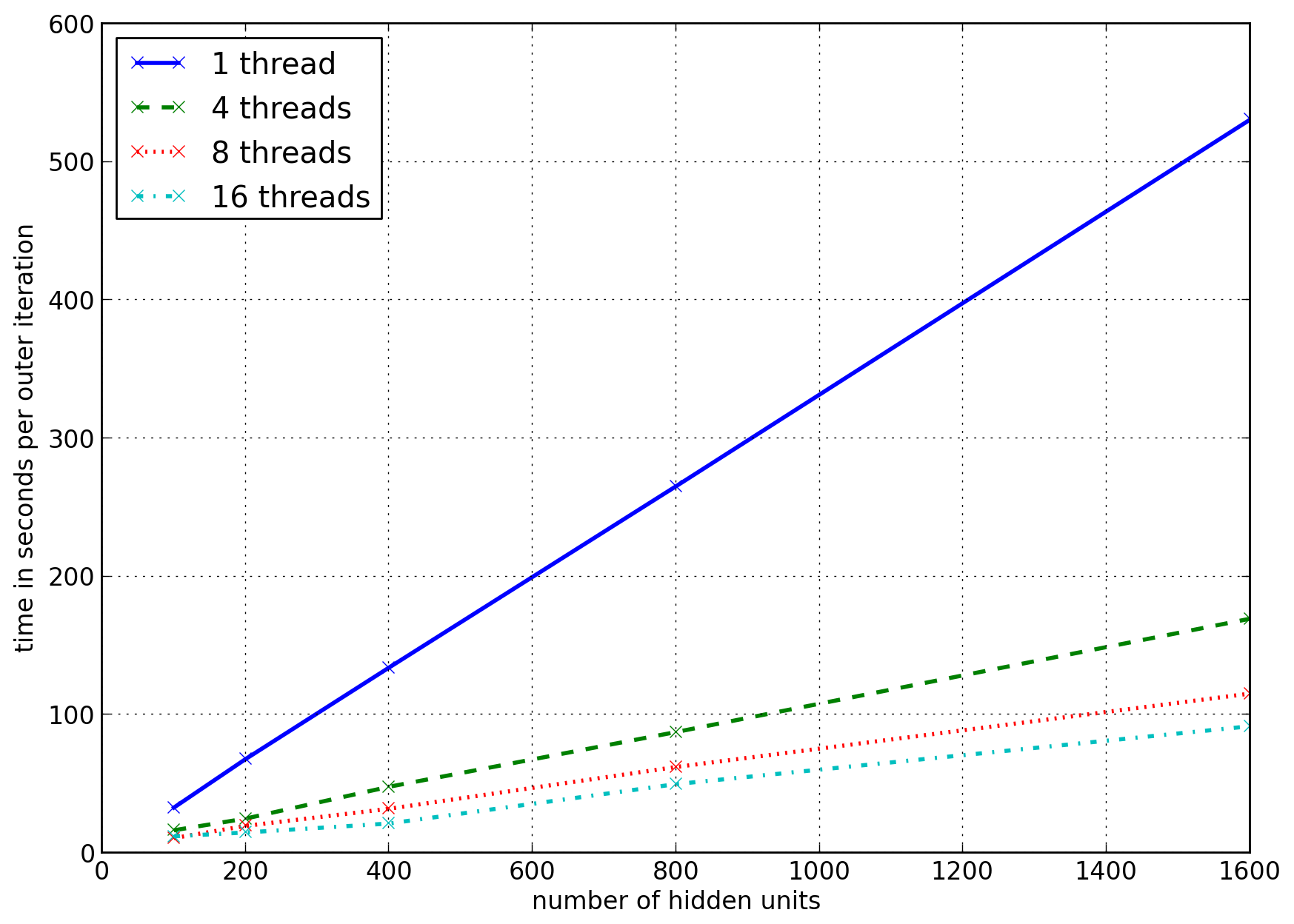}
  }
  \caption{Comparison of performance versus number of threads and number of hidden units in autoencoder layer for HGA with population size (a) 2 and (b) 50.}
  \label{fig:ga_comparison3}
\end{figure*}

In Fig.~\ref{fig:ga_comparison3}, the performance of HGA versus the number of threads and number of hidden units in autoencoder layers is visualized for both a population size of 50 and the default size of 2. As we can see, the HGA's performance scales linearly with the number of hidden units and that this relationship holds even when the population size is increased to 50. Interestingly, the performance when using 8 or 16 threads is not distinguishable for small numbers of hidden units. This is probably due to the computational overhead that result from creating additional threads.

\begin{figure*}[h]
  \centering
  \subfloat[]{
    \includegraphics[width=0.45\linewidth]{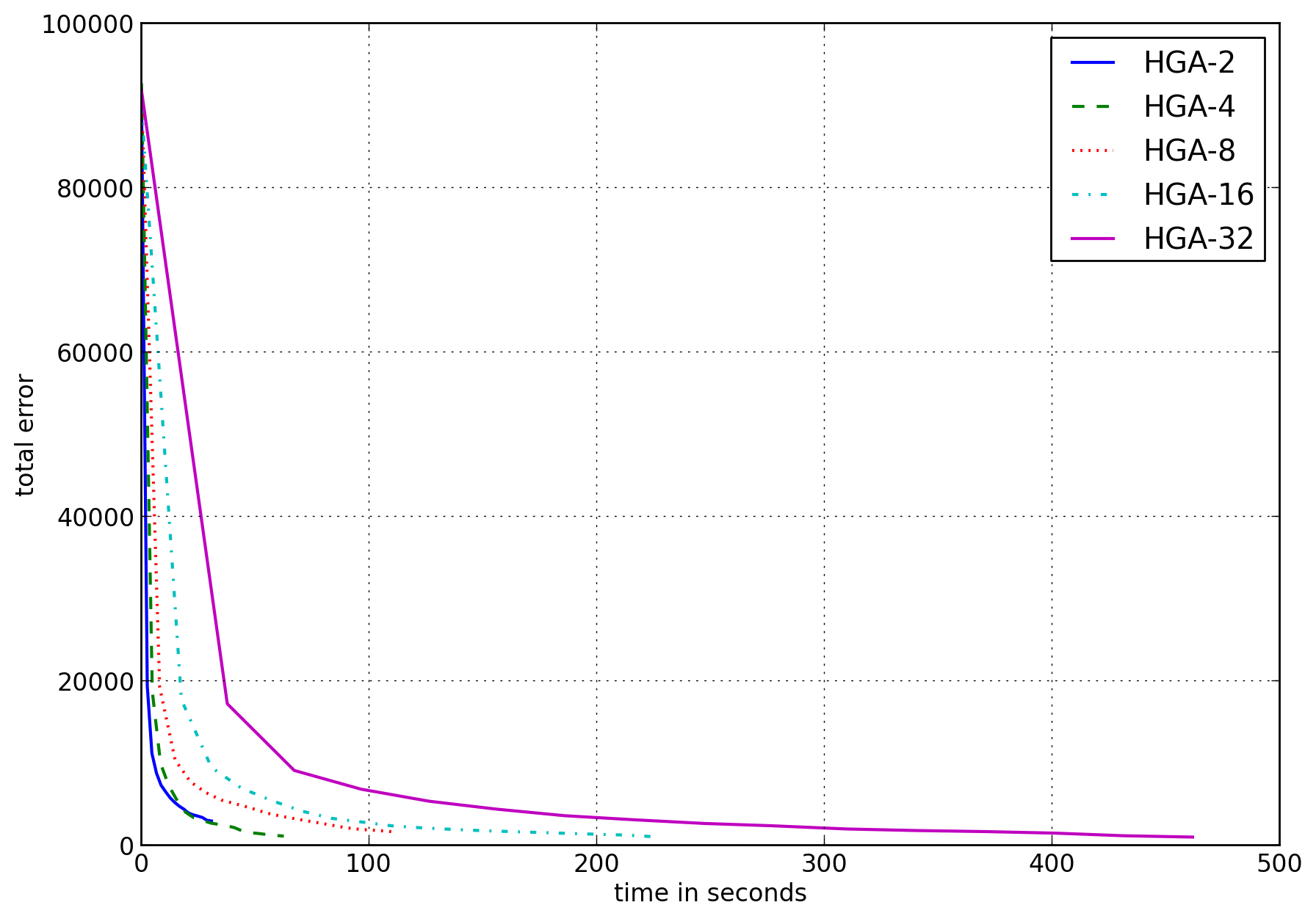}
  }
  \subfloat[]{
    \includegraphics[width=0.45\linewidth]{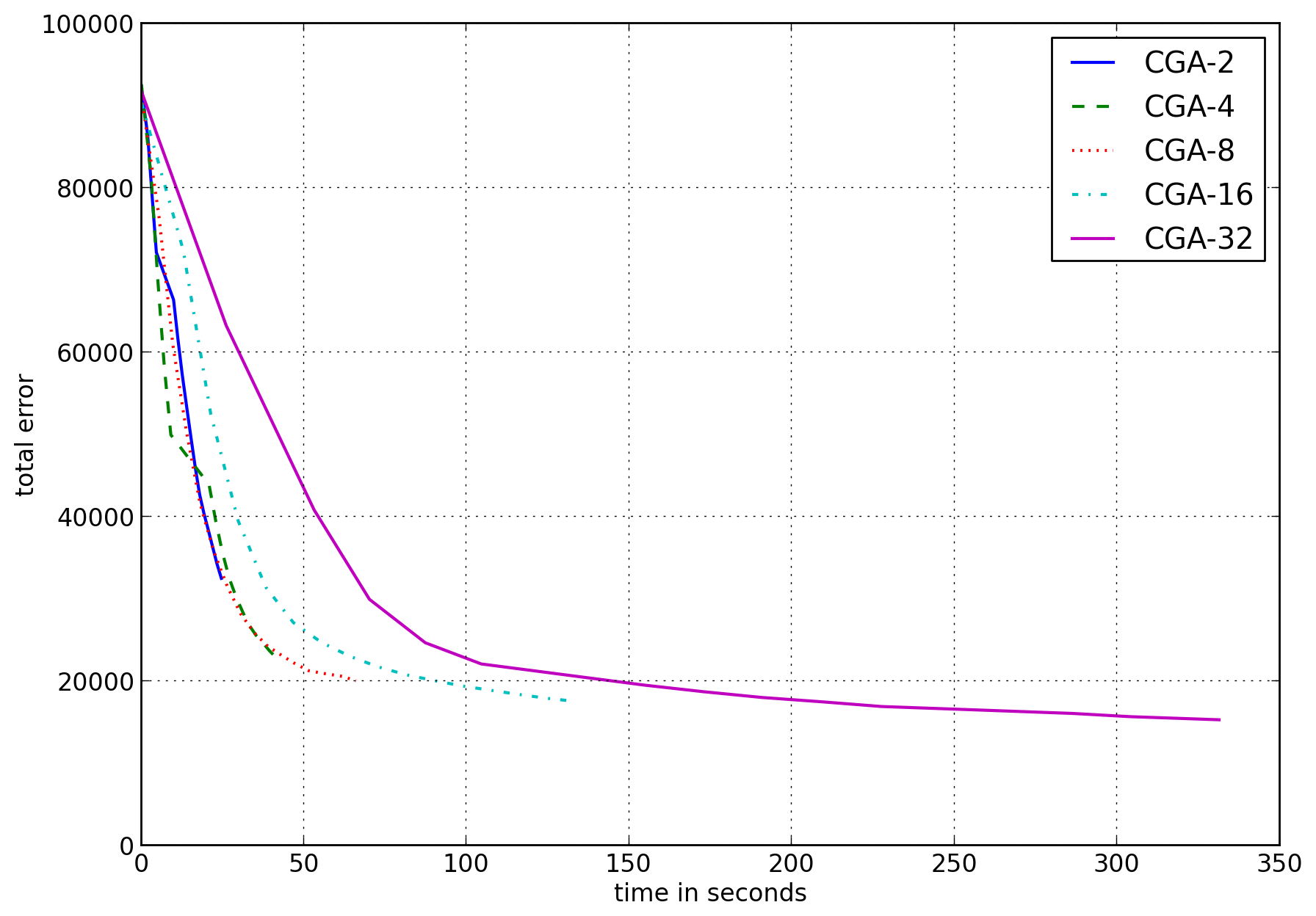}
  }
  \caption{Comparison of performance for (a) HGA and (b) CGA with population sizes of 2, 4, 8, 16, and 32. HGA performs significantly better than CGA in terms of final reconstruction error.}
  \label{fig:ga_comparison4}
\end{figure*}

Finally, in Fig.~\ref{fig:ga_comparison4}, we plot the performance of CGA and HGA for population sizes of 2, 4, 8, 16, and 32. All other hyperparameters remain the same for both HGA and CGA. As we can see, larger populations does lead to a noticeable improvement in the final reconstruction error, but at a cost of increased running time that is linear in the population size. For each fixed population size, HGA performs significantly better than CGA, with a final construction error is roughly three to five times lower. However for a given population size, HGA does take 40\% to 50\% longer to finish; this is attributable to the fact that CGA does not perform backpropagation, which takes additional time. 

% \FloatBarrier
\section{Discussion}
The experimental results demonstrate the importance of learning a better representation of the data for the classification algorithms given. While the kernel SVM performed comparably with and without the improved representation of the data, it is also much more expensive than the linear SVM or logistic regression methods. Thus the trade off becomes whether to use a more expensive classifier, or to learn a better representation and then use a cheaper classifier. It is also important to consider that the autoencoder combined with a simple classifier performed better than the kernel SVM in several cases.

As we saw in the results section, SGD is the fastest method for training the autoencoder, in terms of wall-time. That is, to achieve a given error rate, using SGD will be faster than HGA or CGA in terms of time (as opposed to number of iterations). This might suggest that SGD is the preferred method for training an autoencoder, but it does not come without drawbacks. In terms of the number of iterations to achieve a given error rate, we saw that HGA performed the best, with SGD second and CGA last. This is important because of how the methods scale. SGD scales decently with respect to increasing number of threads, but the genetic algorithms demonstrate better scaling properties, being closer to linear in their scaling. 
Furthermore, HGA is able to achieve the lowest error rate of any of the methods considered. Reconstruction error is an indication of of how well the autoencoder encodes (and decodes) the input; by this metric HGA is a better optimization method than SGD.

However, genetic algorithms have many parameters that need to be tuned and are problem dependent. In contrast, SGD has only one: the learning rate. Consequently it can sometimes be difficult to get GAs to converge at all or at a reasonable rate without knowing the correct parameters. Thus it may be interesting to consider a meta-optimization of the parameters for a genetic algorithm, so that this tuning may be done automatically.

% \FloatBarrier
\section{Future Work}
As mentioned above, genetic algorithms are sensitive to their parameters. Consequently, much time and effort may be required in order to determine the optimal parameters and achieve good convergence performance. Without the correct parameters, the genetic algorithm may not converge to a solution at all or may converge very slowly. A technique known as meta-optimization could be used to automatically determine the correct parameters for performing the optimization. Grefenstette discusses this meta-optimization over six possible parameters for the genetic algorithm \cite{metaoptimization}. Finding the best possible parameters may result in better convergence for our genetic algorithms in both rate of error decrease and minimum error achieved.

Another direction to consider is the adaptive selection of crossover or mutation probability during the training of the autoencoder. This so-called adaptive genetic algorithm reaches the global optimum for a cost function much more quickly than a standard genetic algorithm \cite{srinivas94adaptive}. This adaptive method could be applied to either CGA or to HGA and further improve it's performance.

Our experiments and results are based on shared memory, single machine parallelism. An interesting future topic to explore is expanding the methods we considered to more cores and more machines. Expanding to more cores is a trivial extension requiring just the appropriate multi-core hardware with more than 16 processor cores on a node. Expanding what we have done to a distributed setting would require more work, and rewriting much of the code. Stochastic gradient descent has been shown to be a scalable method for a distributed setting (see, for example \cite{zinkevich2010psgd}) so it seems reasonable that it should scale in this application as well. Genetic algorithms have also been shown to demonstrate good scaling properties for distributed computations \cite{Belding95thedistributed}.

% \FloatBarrier
\section{Conclusion}
We have implemented stacked denoising autoencoders and shown that it achieves accuracy comparable to state of the art classifiers like a SVM with RBF kernel. We also have shown that our autoencoder layers are learning good representations and are capable of denoising and reconstructing the input with little error. These learned representations improve the ability of other classification algorithms to correctly classify the data. We find that SGD scales well with increasing number of hidden units and with increasing number of threads. Lastly, we discovered that HGA, a genetic algorithm that makes use of gradient information, is competitive with SGD, and scales just as well if not better with the size of the autoencoder and number of threads. 

\FloatBarrier

\bibliographystyle{unsrt}
\bibliography{autoencoder}

\begin{thebibliography}{10}

\bibitem{bengio2012rep}
Yoshua Bengio, Aaron~C. Courville, and Pascal Vincent.
\newblock Unsupervised feature learning and deep learning: {A} review and new
  perspectives.
\newblock {\em CoRR}, abs/1206.5538, 2012.

\bibitem{vincent2010stacked}
Pascal Vincent, Hugo Larochelle, Isabelle Lajoie, Yoshua Bengio, and
  Pierre-Antoine Manzagol.
\newblock Stacked denoising autoencoders: Learning useful representations in a
  deep network with a local denoising criterion.
\newblock {\em The Journal of Machine Learning Research}, 11:3371--3408, 2010.

\bibitem{wold1987principal}
Svante Wold, Kim Esbensen, and Paul Geladi.
\newblock Principal component analysis.
\newblock {\em Chemometrics and intelligent laboratory systems}, 2(1):37--52,
  1987.

\bibitem{scholkopf1997kernel}
Bernhard Sch{\"o}lkopf, Alexander Smola, and Klaus-Robert M{\"u}ller.
\newblock Kernel principal component analysis.
\newblock In {\em Artificial Neural Networks—ICANN'97}, pages 583--588.
  Springer, 1997.

\bibitem{tenenbaum2000global}
Joshua~B Tenenbaum, Vin De~Silva, and John~C Langford.
\newblock A global geometric framework for nonlinear dimensionality reduction.
\newblock {\em Science}, 290(5500):2319--2323, 2000.

\bibitem{van2008visualizing}
Laurens Van~der Maaten and Geoffrey Hinton.
\newblock Visualizing data using t-sne.
\newblock {\em Journal of Machine Learning Research}, 9(2579-2605):85, 2008.

\bibitem{hinton2006reducing}
Geoffrey~E Hinton and Ruslan~R Salakhutdinov.
\newblock Reducing the dimensionality of data with neural networks.
\newblock {\em Science}, 313(5786):504--507, 2006.

\bibitem{haykin2004comprehensive}
Simon Haykin.
\newblock Neural networks: A comprehensive foundation.
\newblock {\em Neural Networks}, 2(2004), 2004.

\bibitem{hecht1989theory}
Robert Hecht-Nielsen.
\newblock Theory of the backpropagation neural network.
\newblock In {\em Neural Networks, 1989. IJCNN., International Joint Conference
  on}, pages 593--605. IEEE, 1989.

\bibitem{bottou-91c}
{L\'eon} Bottou.
\newblock Stochastic gradient learning in neural networks.
\newblock In {\em Proceedings of Neuro-N\^imes 91}, Nimes, France, 1991. EC2.

\bibitem{rifai2011contractive}
Salah Rifai, Pascal Vincent, Xavier Muller, Xavier Glorot, and Yoshua Bengio.
\newblock Contractive auto-encoders: Explicit invariance during feature
  extraction.
\newblock In {\em Proceedings of the 28th International Conference on Machine
  Learning (ICML-11)}, pages 833--840, 2011.

\bibitem{bengio2009learning}
Yoshua Bengio.
\newblock Learning deep architectures for ai.
\newblock {\em Foundations and trends{\textregistered} in Machine Learning},
  2(1):1--127, 2009.

\bibitem{schmidhuber2014deep}
J{\"u}rgen Schmidhuber.
\newblock Deep learning in neural networks: An overview.
\newblock {\em arXiv preprint arXiv:1404.7828}, 2014.

\bibitem{hinton2006fast}
Geoffrey Hinton, Simon Osindero, and Yee-Whye Teh.
\newblock A fast learning algorithm for deep belief nets.
\newblock {\em Neural computation}, 18(7):1527--1554, 2006.

\bibitem{lecun1995convolutional}
Yann LeCun and Yoshua Bengio.
\newblock Convolutional networks for images, speech, and time series.
\newblock {\em The handbook of brain theory and neural networks}, 3361, 1995.

\bibitem{hinton2006learning}
Geoffrey~E. Hinton, Simon Osindero, and Yee-Whye Teh.
\newblock A fast learning algorithm for deep belief nets.
\newblock {\em Neural Comput.}, 18(7):1527--1554, July 2006.

\bibitem{ciresan2011committee}
Dan Ciresan, Ueli Meier, Jonathan Masci, and J{\"u}rgen Schmidhuber.
\newblock A committee of neural networks for traffic sign classification.
\newblock In {\em Neural Networks (IJCNN), The 2011 International Joint
  Conference on}, pages 1918--1921. IEEE, 2011.

\bibitem{krizhevsky2012imagenet}
Alex Krizhevsky, Ilya Sutskever, and Geoffrey~E Hinton.
\newblock Imagenet classification with deep convolutional neural networks.
\newblock In {\em Advances in neural information processing systems}, pages
  1097--1105, 2012.

\bibitem{goodfellow2013multi}
Ian~J Goodfellow, Yaroslav Bulatov, Julian Ibarz, Sacha Arnoud, and Vinay Shet.
\newblock Multi-digit number recognition from street view imagery using deep
  convolutional neural networks.
\newblock {\em arXiv preprint arXiv:1312.6082}, 2013.

\bibitem{srinivas1994genetic}
Mandavilli Srinivas and Lalit~M Patnaik.
\newblock Genetic algorithms: A survey.
\newblock {\em Computer}, 27(6):17--26, 1994.

\bibitem{gomez2006efficient}
Faustino Gomez, J{\"u}rgen Schmidhuber, and Risto Miikkulainen.
\newblock Efficient non-linear control through neuroevolution.
\newblock In {\em Machine Learning: ECML 2006}, pages 654--662. Springer, 2006.

\bibitem{floreano2008neuroevolution}
Dario Floreano, Peter D{\"u}rr, and Claudio Mattiussi.
\newblock Neuroevolution: from architectures to learning.
\newblock {\em Evolutionary Intelligence}, 1(1):47--62, 2008.

\bibitem{koutnik2014evolving}
Jan Koutn{\'\i}k, Juergen Schmidhuber, and Faustino Gomez.
\newblock Evolving deep unsupervised convolutional networks for vision-based
  reinforcement learning.
\newblock In {\em Proceedings of the 2014 conference on Genetic and
  evolutionary computation}, pages 541--548. ACM, 2014.

\bibitem{david2014genetic}
Omid~E David and Iddo Greental.
\newblock Genetic algorithms for evolving deep neural networks.
\newblock In {\em Proceedings of the 2014 conference companion on Genetic and
  evolutionary computation companion}, pages 1451--1452. ACM, 2014.

\bibitem{cantu1998survey}
Erick Cant{\'u}-Paz.
\newblock A survey of parallel genetic algorithms.
\newblock {\em Calculateurs paralleles, reseaux et systems repartis},
  10(2):141--171, 1998.

\bibitem{fan2008liblinear}
Rong-En Fan, Kai-Wei Chang, Cho-Jui Hsieh, Xiang-Rui Wang, and Chih-Jen Lin.
\newblock Liblinear: A library for large linear classification.
\newblock {\em The Journal of Machine Learning Research}, 9:1871--1874, 2008.

\bibitem{chang2011libsvm}
Chih-Chung Chang and Chih-Jen Lin.
\newblock Libsvm: a library for support vector machines.
\newblock {\em ACM Transactions on Intelligent Systems and Technology (TIST)},
  2(3):27, 2011.

\bibitem{metaoptimization}
J.J. Grefenstette.
\newblock Optimization of control parameters for genetic algorithms.
\newblock {\em Systems, Man and Cybernetics, IEEE Transactions on},
  16(1):122--128, Jan 1986.

\bibitem{srinivas94adaptive}
M.~Srinivas and L.M. Patnaik.
\newblock Adaptive probabilities of crossover and mutation in genetic
  algorithms.
\newblock {\em Systems, Man and Cybernetics, IEEE Transactions on},
  24(4):656--667, Apr 1994.

\bibitem{zinkevich2010psgd}
Martin Zinkevich, Markus Weimer, Lihong Li, and Alex~J. Smola.
\newblock Parallelized stochastic gradient descent.
\newblock In J.D. Lafferty, C.K.I. Williams, J.~Shawe-Taylor, R.S. Zemel, and
  A.~Culotta, editors, {\em Advances in Neural Information Processing Systems
  23}, pages 2595--2603. Curran Associates, Inc., 2010.

\bibitem{Belding95thedistributed}
Theodore~C. Belding.
\newblock The distributed genetic algorithm revisited.
\newblock In {\em Proceedings of the Sixth International Conference on Genetic
  Algorithms}, pages 114--121, 1995.

\end{thebibliography}

\end{document}